\documentclass[preprint,12pt]{elsarticle}

\usepackage{amssymb}
\usepackage{siunitx}
\usepackage{times}
\usepackage{url}
\usepackage{epsfig}
\usepackage{caption}
\usepackage{graphicx}
\usepackage{amsmath,amssymb} 
\usepackage{mathtools}
\usepackage{booktabs}
\usepackage{amsfonts}
\usepackage{upgreek}
\usepackage[pagebackref=true,breaklinks=true,letterpaper=true,colorlinks,bookmarks=false]{hyperref}
\usepackage[usenames,dvipsnames,svgnames,table]{xcolor}

\usepackage{hyperref}  
\usepackage{capt-of}
\usepackage{dirtytalk}
\usepackage{algorithm}% http://ctan.org/pkg/algorithm
\usepackage[noend]{algpseudocode}
\usepackage[super]{nth}

\usepackage{multirow}
\usepackage{subfigure}
\usepackage[table]{xcolor}
\usepackage{bm}
\usepackage{subfigure}
\usepackage{balance}
\usepackage{hhline,colortbl}
\usepackage{makecell}
\usepackage{comment}
\usepackage{fancyhdr} 
\definecolor{Gray}{gray}{0.9}
\definecolor{Gray1}{gray}{0.97}
\graphicspath{{./}{./figures/}}
\DeclareGraphicsExtensions{.pdf,.jpeg,.png,.jpg,.eps}

\newtheorem{remark}{Remark}

\definecolor{ao}{rgb}{0.0, 0.5, 0.0}

\def\onedot{\ifx\@let@token.\else.\null\fi\xspace}

\makeatother

\def\Vec#1{{\boldsymbol{#1}}}
\def\Mat#1{{\boldsymbol{#1}}}

\newcommand{\RNum}[1]{\lowercase\expandafter{\romannumeral #1\relax}}

\newlength{\Oldarrayrulewidth}

\DeclareMathOperator*{\argmin}{arg\,min}
\usepackage{float}

\journal{Image and Vision Computing}

\begin{document}

\begin{frontmatter}



\title{Cross-Correlated Attention Networks for Person Re-Identification}


% \author{}
\author[1]{Jieming Zhou\corref{cor1}\fnref{fn1}}
\ead{Jieming.Zhou@anu.edu.au}
\author[1]{Soumava Kumar Roy\fnref{fn1}}
\ead{Soumava.KumarRoy@anu.edu.au}
\author[1]{Pengfei Fang}
\ead{Pengfei.Fang@anu.edu.au}
\author[4]{Mehrtash Harandi}
\ead{mehrtash.harandi@monash.edu}
\author[5]{Lars Petersson}
\ead{lars.petersson@data61.csiro.au}

\cortext[cor1]{Corresponding author}

\fntext[fn1]{Equal contribution.}

\address[1]{Australian National University, Canberra, Acton 2601, Australia}
\address[4]{Monash University, Wellington Rd, Clayton VIC 3800, Australia}
\address[5]{Data61, CSIRO, Canberra, Acton 2601, Australia}

% \begin{abstract}
\begin{abstract}
Deep neural networks need to make robust inference in the presence of occlusion, background clutter, pose and viewpoint variations -to name a few- when the task of person re-identification is considered. Attention mechanisms have recently proven to be successful in handling the aforementioned challenges to some degree. However previous designs fail to capture inherent inter-dependencies between the attended features; leading to restricted interactions between the attention blocks. In this paper, we propose a new attention module called Cross-Correlated Attention (\emph{CCA}); which aims to overcome such limitations by maximizing the information gain between different attended regions. Moreover, we also propose a novel deep network that makes use of different attention mechanisms to learn robust and discriminative representations of person images. The resulting model is called the Cross-Correlated Attention Network (\emph{CCAN}). Extensive experiments demonstrate that the \emph{CCAN} comfortably outperforms current state-of-the-art algorithms by a tangible margin.
\end{abstract}

% \end{abstract}

% \includegraphics[width=\linewidth]{Figures_NEW/Structure1_GA.pdf}


% \begin{highlights}
% \item Modeling the inherentspatial relations between different attended regions within the deep architecture.
% \item Joint end-to-end cross correlated attention and representational learning.
% \item State-of-the-art results in terms of mAP and Rank-1 accuracies across several challenging datasets.
% \end{highlights}


\end{frontmatter}


\section{Introduction}
In this paper, we propose a Cross-Correlated Attention Network (CCAN) to jointly learn a holistic attention selection mechanism along with discriminative feature representations for person \underline{Re}-\underline{ID}entification (Re-ID). To this end, we make use of complementary attentional information along a global and a local branch (or feature extractor), in order to localize and focus on the discriminative regions of the input image. 

Person Re-ID refers to the task of judging whether two images, depicting people, belong to the same individual or not. In general, the two images are obtained from two distinct cameras without any overlapping views. More specifically, given a query image containing the person of interest (or probe), Re-ID aims to find all the images that contain the same identity (id) , as that of the query image, from a large gallery set \cite{zheng2016person}.

Any robust Re-ID algorithm is required to address the following challenges: (1) viewpoint variations in visual appearance and environmental conditions due to different non-overlapping camera views, (2) significant pose changes for the same probe across time, space and camera views, (3) background clutter and occlusions, (4) different individuals may have similar appearance across different cameras or vice versa, (5) low resolution of the images limiting the use of face based biometric systems~\cite{decann2015modelling}. All these factors lead to significant visual deformations across the multiple camera views for the same person of interest.

In order to overcome these challenges, most of the early works focused on (1) designing discriminative \emph{hand-engineered} feature representations which are invariant to lighting, pose and viewpoint changes, and occlusion or clutter~\cite{zheng2016person,farenzena2010person}; (2) learning a robust \emph{distance metric} for similarity measurement such that the embedded feature vectors belonging to the same class are closer to each other compared to the ones from different classes~\cite{chen2016similarity, koestinger2012large}. 

With the success of Deep Learning (DL) algorithms~\cite{lecun2015deep} across a large number of tasks in computer vision, recent deep Re-ID algorithms combine both the aforementioned aspects together into a unified end-to-end framework. While some deep algorithms address Re-ID by developing distinct global feature extraction units~\cite{AhmedEjaz2015CVPRAnimprovedDeepLearningArchitectureforReID,li2014deepreid},
others use a hybrid model which holistically combines the global and local features for an improved performance~\cite{TianMaoqing2018CVPREliminatingBackgroundBiasforReID,LiDeepJoint}.
Body-part detectors have been pre-dominantly used to extract local features that are distinct, discriminative and compatible with global features~\cite{LiDangwei2017CVPRLearningDeepContextAwarefeaturesforReID,LiWei2018CVPRHarmoniousAttentionNetforReID}.
Similarly, pose estimation, correction and normalization networks~\cite{su2017pose,ZhengLiang2017arXivPoseInvariantEmbeddingforReID, Sarfraz_2018_CVPR}
have also shown great potential with, or without, part detectors in handling misalignment and viewpoint variations prevalent in the Re-ID datasets. The use of such special purpose auxiliary information tend to improve upon the methods it is applied to.

\emph{Attention} based person Re-ID models have also been showing promising results as of late. Attention, as the name suggests, is comprised of two basic conceptual functionalities: ``\emph{where to look}'' and ``\emph{how carefully to look}''. \emph{Hard-attention} often uses a window produced by, \emph{e.g},  a Spatial Transformer Network (STN)~\cite{JaderbergMax2015NIPSSTN} that models the former with a binary mask over the input features, whereas \emph{soft-attention} simulates the latter by importance weighting of the input features~\cite{xu2015show}. 

Both these attention based learning approaches have been successfully integrated when addressing the person Re-ID task~\cite{LiDangwei2017CVPRLearningDeepContextAwarefeaturesforReID,LiWei2018CVPRHarmoniousAttentionNetforReID}. However, these models do not capture spatial inter-dependencies (\emph{i.e}, \emph{self-attention}) within the input features, thereby failing to recognize and perceive spatially distant, yet visually similar regions. They also do not capture (or improve) any inter- (or cross-correlated) dependencies between the separately attended regions, thus failing to boost the overall Signal-to-Noise Ratio (SNR) in the learnt feature maps. Moreover, convolutional based soft-attention blocks are not able to capture the inherent contextual information that exist in the input features.

To address the aforementioned drawbacks, we design the CCAN, a novel yet intuitive \emph{Cross-Correlated} Attention based deep network. CCAN consists of a novel attention module which aims to \emph{exploit} and \emph{explore} the correlation between different regions at various levels of a deep model. It also benefits from a top-down interaction scheme between the global and local feature extractors through the different attention modules to automatically focus and extract distinct regions in the input image for enhanced feature representation learning. 

The major contributions of our work are as follows:
\begin{itemize}
  
  \item A novel \emph{Cross-Correlated} Attention (CCA) module to model the inherent spatial relations between different attended regions within the deep architecture.
  
  \item A novel deep architecture for joint \emph{end-to-end} cross correlated attention and representational learning.

  \item State-of-the-art results in terms of mAP and Rank-1 accuracies across several challenging datasets such as Market-1501~\cite{zheng2015scalable} and DukeMTMC-reID \cite{ristani2016MTMC}, CUHK03~\cite{li2014deepreid} and MSMT17~\cite{wei2018person}.

\end{itemize}

% \vspace{-0.1cm}
\section{Related Work}\label{RelatedWork}
% \vspace{-0.1cm}
Much of the earlier work in person Re-ID was focused on hand-engineered feature representations~\cite{liao2015person,li2016richly,wang2016highly,zhong2017re,zheng2016person} 
or learning a robust metric~\cite{zheng2013reidentification,koestinger2012large,xiong2014person} 
to overcome the associated challenges. % of person Re-ID. 
Recent studies employ Deep Neural Networks (DNNs) for joint learning of the discriminative features and similarity measures in end-to-end frameworks~\cite{AhmedEjaz2015CVPRAnimprovedDeepLearningArchitectureforReID,ChengDe2016CVPRPersonReIDbyMultiChannelPartswithImprovedTriplet}. Since we are chiefly interested in attention methods for person Re-ID in this paper, we will not cover part/pose-based solutions here and refer interested readers to~\cite{su2017pose,ZhengLiang2017arXivPoseInvariantEmbeddingforReID,sun2018beyond}.

To address the viewpoint/pose variations and misalignment issues commonly present in a Re-ID system, a profound idea is to benefit from the use of attention techniques in DNNs~\cite{zhao2017deeply,LiWei2018CVPRHarmoniousAttentionNetforReID,LiuXihui2017ICCVHydraPlusNetforReID,LiuHao2017arXivEndtoEndComparativeAttentionNetworksforReID,LiDangwei2017CVPRLearningDeepContextAwarefeaturesforReID,Mancs,AANet_A,Fang_2019_ICCV}. Li~\emph{et al.}~\cite{LiDangwei2017CVPRLearningDeepContextAwarefeaturesforReID} used a Spatial Transformer Network (STN)~\cite{JaderbergMax2015NIPSSTN} as a basis for creating a form of \emph{hard-attention} to search and focus on the discriminative regions in the image, subject to a pre-defined spatial constraint. Zhao~\emph{et al.}~\cite{zhao2017deeply} designed a novel hard-attention module (with components similar to STN) and integrated it into a CNN. This helped to focus on more discriminative regions. Subsequently, by extracting and processing features from the attention regions, improvements to the overall performance were observed. AANet \cite{AANet_A} proposed a Part Feature Network by cropping body parts according to the location of the peak activation in the feature maps. Arguably, hard-attention modules fail to capture the coherence between image pixels within the attention windows due to their inflexible modelling nature. The Comparative Attention Network (CAN)~\cite{LiuHao2017arXivEndtoEndComparativeAttentionNetworksforReID} employs LSTMs to perform soft-attention at a holistic scale and identify discriminative regions in Re-ID images. Liu \emph{et al.} \cite{LiuXihui2017ICCVHydraPlusNetforReID} proposed \emph{HydraPlus-Net} (HPN) which utilizes soft-attention across multiple scales and levels to learn discriminative representations. Dual ATtention Matching networks (DuATMs)~\cite{DuATM} use spatial bi-directional attentions along sequence matching to learn context-aware feature representations. Wang~\emph{et al.} proposed \emph{Mancs}~\cite{Mancs} and designed a soft-attentional block and a novel curriculum sampling method to learn focused attention masks. In contrast to the aforementioned algorithms, HA-CNN \cite{LiWei2018CVPRHarmoniousAttentionNetforReID} uses both hard and soft attention modules to efficiently learn \say{\emph{where to look}} and \say{\emph{how carefully to look}} simultaneously.

Recently, Zhou \emph{et al.}~\cite{zhou2019discriminative} propose a novel attention regularizer along with a novel triplet loss which consistently learns correlated attention masks from low, mid and higher level feature maps within an interactive loop. 
DGNet \cite{zheng2019joint} proposed coupling person re-id learning and image generation in a unified joint learning framework such that the re-id learning stage can benefit from the generated data with an inherent feedback loop to learning a superior embedding space. CAMA~\cite{yang2019towards} enhances learning of traditional global representations for person Re-ID by learning class activation maps to discover discriminative and distinct visual features. CASN~\cite{zheng2019re} designed a new siamese framework in order to learn discriminative attention masks and enforce attention consistency among images of the same person. Likewise, OSNet~\cite{zhou2019omni} designed a new aggregation gate that dynamically fuses features at multiple different scales with channel-wise attentional weights. MHAN~\cite{chen2019mixed} proposed the High-Order Attention (HOA) to integrate complex and higher order statistical information in learning an attention mask so as to capture and distinguish subtle differences between the pedestrian and the background.

In contrast to the aforementioned techniques, CCAN makes use of a novel, yet intuitive, \emph{cross-correlated} attention module which discovers and exploits inter-correlated spatial dependencies in the learnt feature maps. It then propagates these learnt dependencies along the feature extraction units to inherently learn robust and discriminative features and attention maps; thereby improving the overall information gain in a data-driven fashion.

\section{Cross-Correlated Attention Networks}
\label{CCAN}

Let $\Mat{x}_i \in \mathcal{X}$  be an image, with  $\mathcal{X} \subset \mathbb{R}^{H \times W \times C}$ denoting the image-space, where $H, W$  and $C$ indicate its rows, columns and channels, respectively. In person Re-ID, we are provided with $N$ pairs of the form $\left \{ \Mat{x}_i, y_i\right \}_{i = 1}^N$ with $y_i \in \{1,\cdots,K\}$ representing the identity of the person depicted in $\Mat{x}_i$. The aim, here, is to learn a generic non-linear mapping $\Psi:\mathcal{X} \to \mathcal{H}$ from the image space $\mathcal{X}$ onto a latent feature space $\mathcal{H}$ such that, 
in $\mathcal{H}$, embeddings coming from the same identity are closer to each other than those of different identities. {We achieve this by exploiting the complementary nature of global and local information in Re-ID images using a combination of two different, and complementary, learnable attention modules. We first provide a detailed overview of the attention modules (\textsection{\ref{sec:attention}}); followed by the overall structure of CCAN (\textsection{\ref{MAN_Overview}}).}

\subsection{Attention Layers}
\label{sec:attention}
In CCAN, we introduce a variation of self-attention named \emph{Cross-Correlated} Attention. The Cross-Correlated Attention mechanism aims to capture, exploit and boost spatial inter-dependencies (or cross-correlation) between different selected regions. 

\begin{figure}[ht]
  \centering
  \includegraphics[width=8.2cm,keepaspectratio]{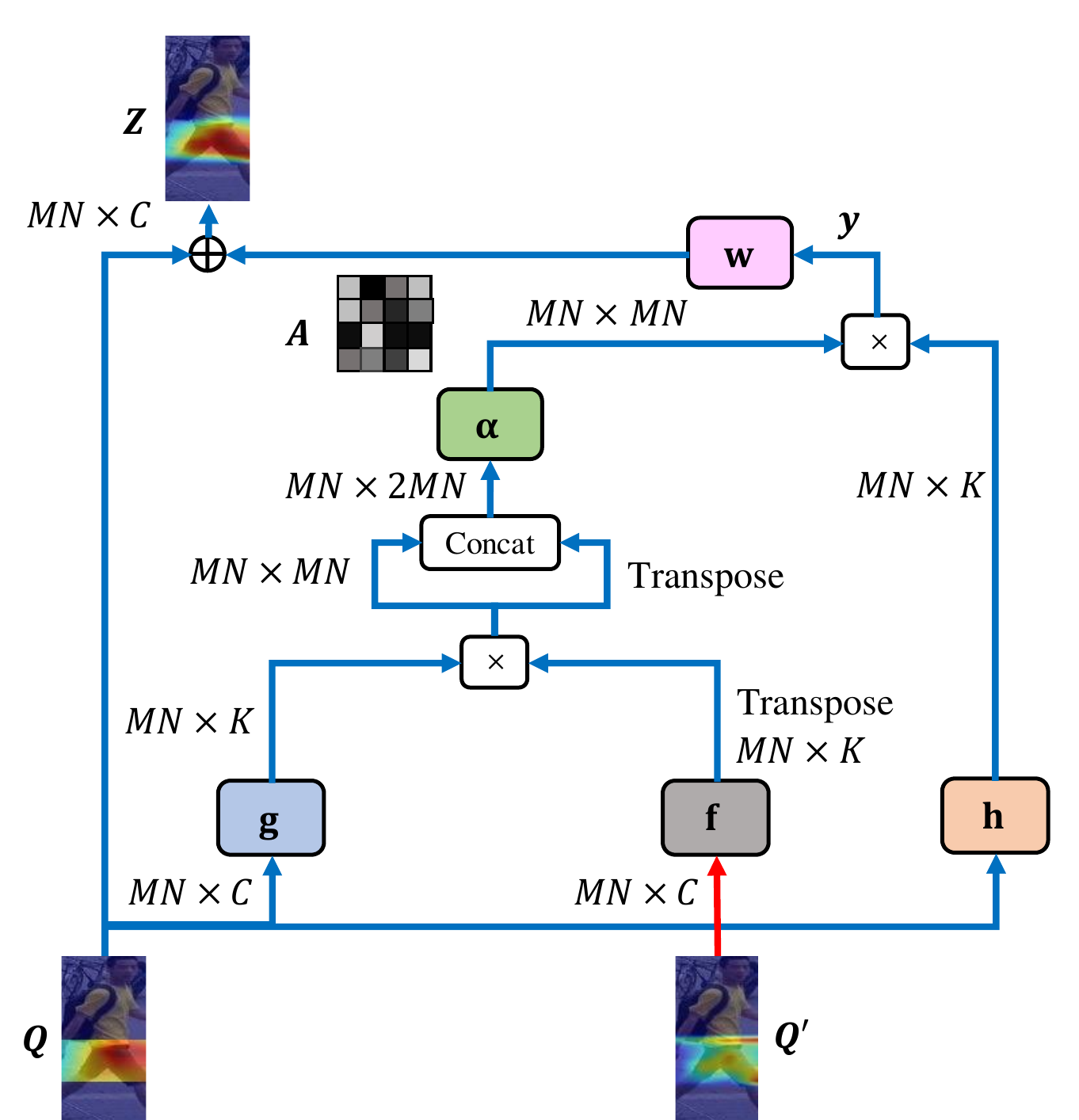}
  \caption{\small{The architecture of Cross-Correlated Attention (CC-Attention) used in our model (blue blocks in Fig.~\ref{fig:CCANstrucutre}). CC-Attention is able to find correlated spatial locations in its two different input feature maps, which are further processed by the subsequent processing layers for discriminative feature learning.}} \label{fig:Sstrucutre}
\end{figure}

The Cross-Correlated Attention (CC-Attention or CCA) module which aims to model the cross-correlation (or inter-dependencies) between different feature maps as a means to construct the attention mask. Each CCA module accepts two inputs and calculates the attention as a weighted combination of the input feature maps (see Fig.~\ref{fig:Sstrucutre} for a conceptual diagram). This, as will be shown empirically, captures the inter-dependencies between the spatial regions in various feature maps with only a small computational overhead.

The CCA block works with the so-called positional matrices 
$\Vec{q},\Mat{q}' \in \mathbb{R}^{MN \times C}$.
In our application, the positional matrices are constructed from two  feature maps $\Mat{Q},\Mat{Q}' \in \mathbb{R}^{M \times N \times C}$ via reshaping through spacial dimension, \emph{i.e} $\mathbb{R}^C \ni \Vec{q}_i   = \Mat{Q}(m,n ,:)~~\forall i \in \{1, \ldots, MN\},\forall m \in \{1, \ldots, M\},\forall n \in \{1, \ldots, N\}$. The matrices $\Mat{q}$ and $\Mat{q}'$ are then transformed into two feature spaces using  independent non-linear mappings \textbf{g} and \textbf{f}, respectively. The non-linear mappings are realized through $\textbf{f}(\Vec{q}'_i) = \phi\big(\Vec{q}'_i~{\Mat{W}_\textbf{f}}\big)$ and $\textbf{g}(\Vec{q}_i) = \phi\big(\Vec{q}_i~{\Mat{W}_\textbf{g}}\big)$, 
where $\Mat{W}_\textbf{f}, \Mat{W}_\textbf{g} \in \mathbb{R}^{K \times C}$, where the non-linearity $\phi:\mathbb{R} \to \mathbb{R}$ acts element-wise on $\textbf{f}$ and $\textbf{g}$. In our experiments, we choose $\phi(x) = \mathrm{ReLU}(x) = \max(0,x)$. These two spaces are then used to calculate a primary attention map between the inputs at the different spatial locations as follows:
\begin{equation}
\label{con:selfAttenMap}
\Mat{A} = \phi\big(\big[\Mat{A}', \Mat{A}'^\top\big]\Mat{W}_{\boldsymbol\upalpha}\big)~~,
\end{equation}
where $\Mat{A}'=\textbf{g}{(\Vec{q})} \textbf{f}{(\Vec{q}')}^\top$, $[~.~,~.~]$ denotes the concatenation operation along the width. Furthermore, $\boldsymbol\upalpha$ is a linear layer with weight ${\Mat{W}_{\boldsymbol\upalpha}} \in \mathbb{R}^{2MN \times MN}$. $[\Mat{A}]_{ij}$ is a measure of spatial dependencies between the $i^{th}$ and the $j^{th}$ spatial locations of the positional matrices $\Mat{q}$ and $\Mat{q}'$ respectively; thereby realizing a measure of cross-correlation between them. The symmetric operation described above guides the CCA module to focus on the correlated positions in both the $\Mat{q}$ and $\Mat{q}'$, which is processed by the subsequent layers of the network. The resultant map $\Mat{A}$ is then used to generate $\Mat{y} \in \mathbb{R}^{MN \times K}$ for input $\Mat{q}$ as follows:
\begin{equation}
    \Vec{y}_i = \frac{1}{MN} \sum_{j=1}^{MN} {\big([\Mat{A}]_{ij} \odot \phi \big(\textbf{h}(\Vec{q}_j)\big)\big)}, ~ \forall i = 1~....~MN,
    \label{con:selfAttenyi}
\end{equation}
where $\odot$ is Hadamard (element-wise) product , $\Vec{y}_i$ is a weighted combination of the responses at all positions denoted by $j$, and $\textbf{h}$ is also a non-linear layer with its weight ${\Mat{W}_\textbf{h}} \in \mathbb{R}^{K \times C}$ such that $\textbf{h}(\Vec{q}_i) = \phi\big({\Vec{q}_i~\Mat{W}_\textbf{h}}\big)$. 
We further pass $\Mat{y} $ through a linear layer $\textbf{w}$ to obtain the final output of the CC-Attention module as follows
\begin{equation}
    \Vec{z}_i  =  \textbf{w}(\Vec{y}_i) + \Vec{q}_i ,~   ~~~ \forall i = 1, \ldots, mn
    \label{con:selfAttenzi}
\end{equation}
with $\Vec{z} =[\Vec{z}_1;\cdots;\Vec{z}_{mn}] \in \mathbb{R}^{MN \times C}$ and $\Vec{z}_i \in \mathbb{R}^C$, %, i \in \{1, \ldots, mn\}$, 
and $\textbf{w} (\Vec{y}_i ) = \phi\big(\Vec{y}_i~{\Mat{W}_\textbf{w}}\big)  $ such that $\Mat{W}_\textbf{w} \in \mathbb{R}^{C\times K}$. The output $\Vec{z}$ is reshaped to $\mathbb{R}^{M \times N \times C}$ to match that of input $\Mat{Q}$. In all our experiments, we have fixed the value of $K$ to be $C/8$.

An intuitive way of thinking about the CCA module is to see $\textbf{g}$ and $\textbf{f}$ as non-linear signatures of elements $\Vec{q}$ and $\Vec{q}'$. The cross-correlation between the non-linear signatures acts as a gate and controls the information flow based on inter-correlation for generating the mask. The information, here, is encoded through $\textbf{h}$. The result is further pruned by $\textbf{w}$ and generates the attention map in an additive form. The additive form resembles the residual computing which is proven to be beneficial in training deep architectures.

\begin{remark}
In the CCA module, we have introduced a symmetric cross-correlation operation between its input feature maps $\Vec{q}$ and $\Vec{q}'$ to generate the attention map $\Mat{A}$ (see Eqn.~\ref{con:selfAttenMap}). It thereby encapsulates symmetrical inter-dependencies between its inputs. The standard cross-correlation operation does not take into account such symmetric relationships between the inputs. We believe that this subtle change makes CCA attend to highly correlated regions in both of its input feature maps. 
\end{remark} 

\begin{figure}[h]
  \centering
  \includegraphics[width=8.2cm,keepaspectratio]{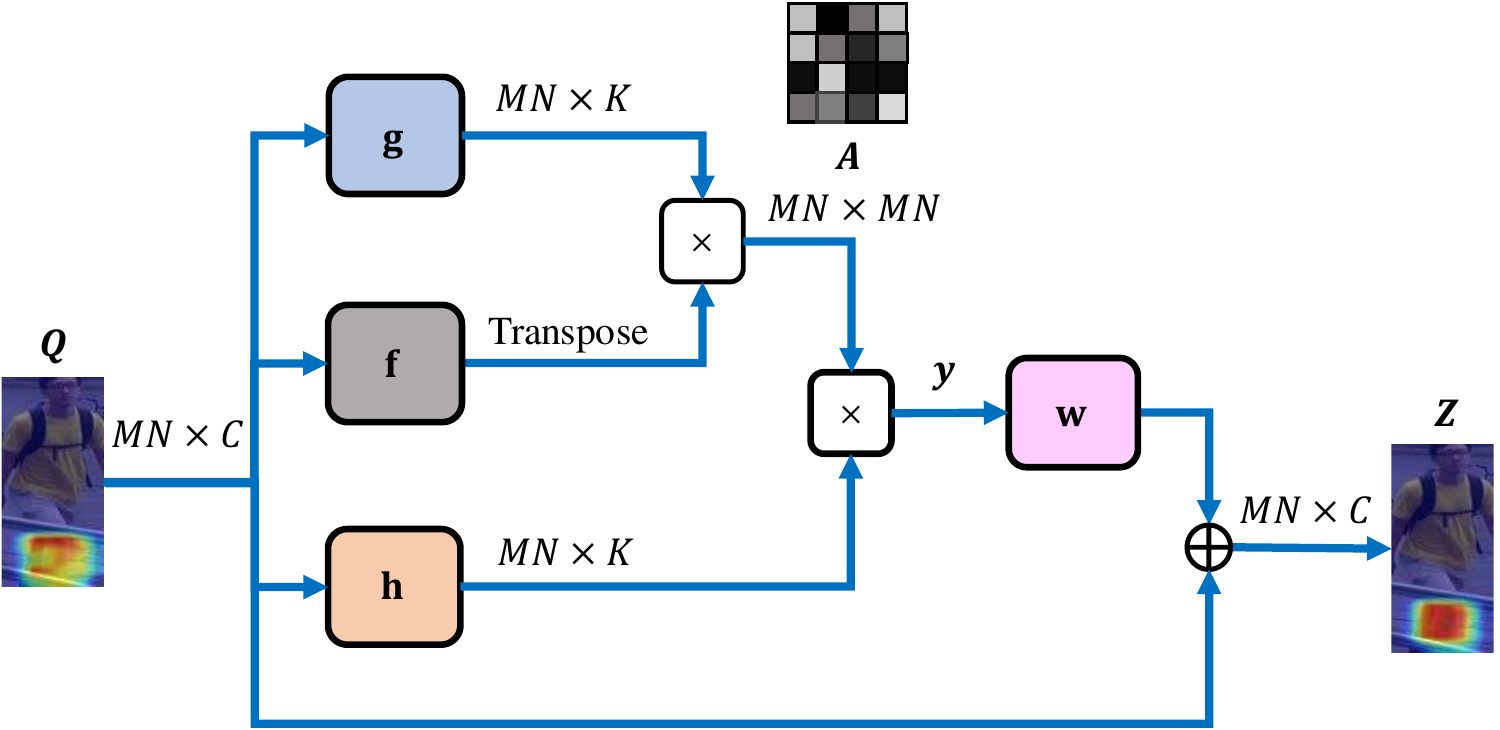}
  \caption{\small{Schematic of the Non-Local Self-Attention module as defined in~\cite{WangXiaolong2017CVPRNonLocalNN}.}} \label{fig:Sstrucutre_original}
\end{figure}

\begin{remark}
When $\Mat{Q}=\Mat{Q}'$, the overall structure represents a form of Symmetric Self-Attention (SS-Attention or SSA) that aims to model highly correlated regions within itself. This form of symmetric self-attention is applied in the global branch, (\emph{i.e}, ${\mathrm{SS}^G_1}$) which models the intra-dependencies within the input. Further simplification of the SS-Attention module by removing the \say{Concat} and \say{$\ \boldsymbol\upalpha$} block leads to the Non-Local Self-Attention module which is shown in Fig.~\ref{fig:Sstrucutre_original}. Thus we equip the traditional self-attention module with these two important changes to model symmetric cross-correlation attention between its two different inputs.
\end{remark}

\begin{figure*}[t]
  \centering
  \includegraphics[width=\linewidth,keepaspectratio]{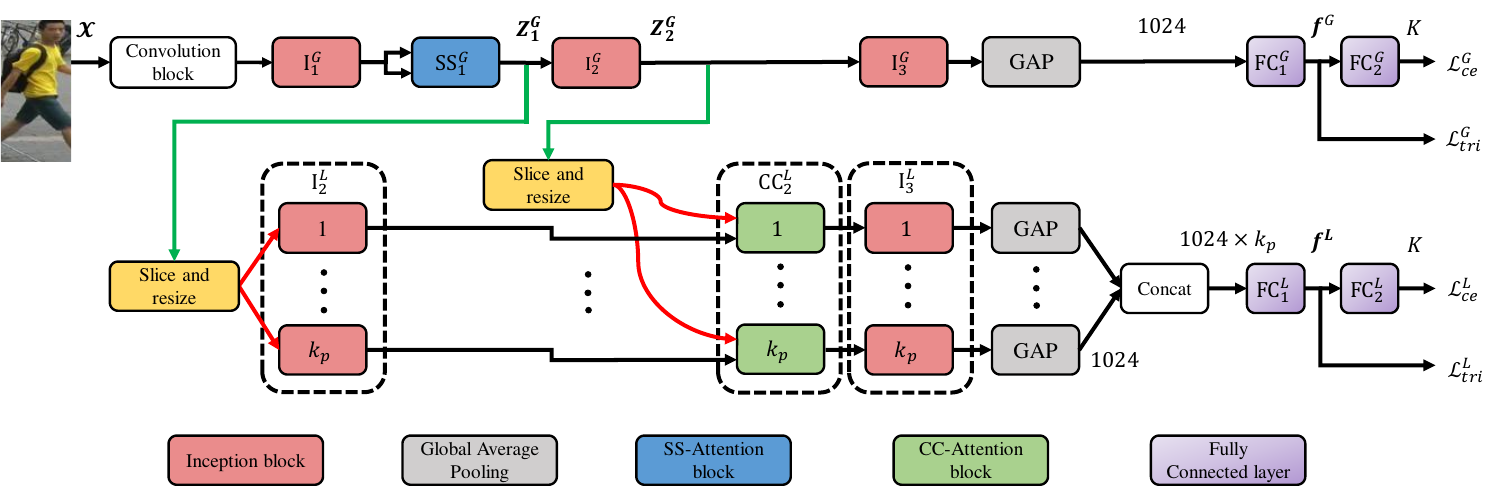}
  \caption{\small{Architecture of CCAN. $G$ and $L$ denote global and local branches. The local branch has $k_p$ sub-branches.
  The local branches receive part patches from the global branch (\emph{i.e} $\Mat{Z}_1^G$ and $\Mat{Z}_2^G$).
  Building blocks of the sub-branches are shown by dashed boxes (refer to~\textsection~\ref{MAN_Overview} for more detail). $\mathcal{L}_{ce}$ and $\mathcal{L}_{tri}$ denote cross-entropy and triplet loss respectively. Green arrows indicate inputs for creating part patches.}}
  \label{fig:CCANstrucutre}
\end{figure*}

\subsection{Structure of the CCAN}
\label{MAN_Overview}
A CCAN consists of two main branches (\emph{i.e}, streams or feature extractors), namely the \emph{global}, $G$, and 
the \emph{local}, $L$, branch (see Fig.~\ref{fig:CCANstrucutre} for an overview of the architecture of CCAN). The purpose of the global branch is to capture and encode the overall appearance of a person, while the local branch encodes part information. The local branch, itself, has $k_p$ sub-branches (or part-streams).

The basic building block of all branches is the \emph{Inception block} of GoogLeNet~\cite{szegedy2015going}. The global branch makes use of three {Inception} blocks, $\left \{ \mathrm{I}_k^{G} \right \}_{k=1}^3$ along with a self-attention module {$\mathrm{SS}_1^{G}$} to encode the global appearance ({$\mathrm{I}_k^{\bullet}$} marks the beginning of the $k$-{th} level of processing in CCAN). The Inception blocks in the global stream enable us to analyze the input at various resolutions, thereby realizing a coarse to fine global representation. The local branch, as the name implies, attends to the local and discriminative parts of the input image. The local branch comprises of $k_p$ sub-branches, each intended to extract features belonging to a distinct part in the input image. For the $s$-th sub-branch, we denote its Inception blocks by $\mathrm{I}^L_{k, s}$ with $k \in \{2,3\}$ and $s \in \{1, \cdots, k_p\}$ (see Fig.~\ref{fig:CCANstrucutre} for details). We emphasize that each $\mathrm{I}_{k,\bullet}^{L}$ is an independent module, meaning that weights are not shared across the $k_p$ part-streams. 

In order to feed part information into local branches, we slice the feature maps at $\Mat{Z}_1^G$ and $\Mat{Z}_2^G$ (\emph{i.e} the input and output of $\mathrm{I}_2^{G}$) into $k_p$ horizontal equal patches independently. Thereafter, all the sliced patches are resized to the size of their corresponding feature maps using bilinear interpolation. Moreover, each of the sub-branches consist of a cross-correlated attention module (\emph{i.e} $CC^L_{2, s}$) $\forall s \in \{1, \cdots, k_p\}$. Every $CC^L_{2, s}$ calculates the cross-correlation between the sliced part patches of $\Mat{Z}_1^G$ (after having been passed through $\mathrm{I}_{k,\bullet}^{L}$) and $\Mat{Z}_2^G$ in each of the sub-branches independently. This sharing of feature mapsbetween the attention modules across the global and local branch within CCAN leads to the discovery of highly correlated regions;
thereby realizing a simple but effective CCA scheme within CCAN. 

The global branch is appended with a \emph{global average pooling} (GAP) layer and two fully connected ($\mathrm{FC}^G_1$ and $\mathrm{FC}^G_2$) layers, with the output of the $\mathrm{FC}^G_1$ realizing a $d$-dimensional embedding space. Similarly, the outputs of local sub-branches are passed through GAP layers and concatenated to produce a $1024 \times k_p$ feature vector. This is then passed through $\mathrm{FC}_1^L$ to produce the $d$-dimensional embedding vector in the local branch, which is further passed through $\mathrm{FC}^L_2$. It should be noted that the  $\mathrm{FC}^G_2$ and $\mathrm{FC}^L_2$ realize representations suitable for classification (\emph{i.e}, $\mathcal{L}_{ce}^G$ and $\mathcal{L}_{ce}^L$). As such, their output dimensionality is $K$, the number of identities in the training set. We will discuss this in more detail later.

\subsection{Loss function}
\label{sec:loss}
Following the common practice in learning embeddings~\cite{weinberger2009distance,oh2016deep,hu2014discriminative,song2017deep}, we make use of a combination of classification and ranking losses (cross entropy loss with \underline{L}abel-\underline{S}moothing \underline{R}egularization (LSR) \cite{Szegedy2016RethinkingTI} and the semi-hard triplet loss~\cite{SchroffFlorian2015CVPRFaceNet,manmatha2017sampling}, respectively), to jointly optimize the global and the local branch. The overall loss is defined as follows:\begin{equation}\label{eq:TotalLoss}
\mathcal{L}_{tot} = \mathcal{L}_{ce}^G + \mathcal{L}_{tri}^G + \mathcal{L}_{ce}^L  + \mathcal{L}_{tri}^L ~~,
\end{equation} where the subscripts \say{$ce$} and \say{$tri$} denote the cross-entropy and triplet loss respectively. Moreover, the superscripts $G$ and $L$ indicate the global and local branch. We briefly describe the semi-hard triplet mining strategy used in our algorithm for calculating the triplet loss.

\subsubsection*{Semi-hard Triplet Mining}

In each mini-batch of $N$ training samples, we \textit{mine} $|P|$ triplets of the form $\left\{\left( \Vec{x}_i^{a}, \Vec{x}_i^{p}, \Vec{x}_i^{n} \right)  \right\}_{i=1}^{|P|}$, with the constraint that $\left(\Vec{x}_i^{a}, \Vec{x}_i^{p}\right)$ are in the same category, while $\left( \Vec{x}_i^{a}, \Vec{x}_i^{n} \right)$ are not. We also use the semi-hard mining strategy~\cite{SchroffFlorian2015CVPRFaceNet} to generate robust triplets for training the network. More specifically, given the anchor $\Vec{x}^{a}$ and its positive example $\Vec{x}^{p}$, we obtain the top $r$ semi-hard negative triplets as follows

\begin{align*}
\Vec{x}^{n} =  \left \{ \Vec{x}^j : \argmin_{D_a^p~<~D_a^j;~ \forall j = 1, \cdots, r} D_a^j~;~ s.t.~D_a^j <D_a^{j+1} \right \}~~,
\end{align*}
\noindent where $D_a^j = \big \| \Vec{x}^a \hspace*{-0.5ex} - \hspace*{-0.5ex} \Vec{x}^j \big \|^2$. $r$ is set to $10$ for all the datasets. Moreover, to avoid any degeneracy, we randomly pick $v$ different identities and sample ${N}/{v}$ random images from each of the selected identities to create the mini-batch. These triplets are then used to compute the triplet embedding loss: 
\begin{align}
\label{eqn:triplet}
	 \mathcal{L}_{tri} = \hspace*{-0.5ex} \frac{1}{|P|} \hspace*{-0.5ex} \sum_{i=1}^{|P|} \hspace*{-0.5ex} 
	\Big[ \big \| \Vec{x}_i^a \hspace*{-0.5ex} - \hspace*{-0.5ex} \Vec{x}_i^p \big \|^2  \hspace*{-1.5ex} - \hspace*{-0.5ex}
	\big \| \Vec{x}_i^a \hspace*{-0.5ex} - \hspace*{-0.5ex} \Vec{x}_i^n \big \|^2 
	\hspace*{-1.0ex} + \hspace*{-0.5ex} \tau \Big]_+ \hspace*{-0.5ex} ~~~,
\end{align}
where $[y]_+ = \max(0,y)$ is the hinge loss, and $\tau > 0$ is a user-specified margin. 

\subsection{Person Re-ID by CCAN}
Given a trained CCAN model and an input image $\Mat{x}_i$; we first obtain its $d$ dimensional global feature $\Mat{f}^G_i$ and $d$ dimensional local feature $\Mat{f}^L_i$. We perform L2 normalization on each of them separately, and then proceed to concatenate them to obtain the joint $2d$ feature vector $\Mat{f}_i^A = \left [ \Mat{f}_i^G; \Mat{f}_i^L  \right ]$. Thus, given a probe image $\Mat{x}_p$ from one camera view and all the gallery images $\left \{ \Mat{\Bar{x}}_{j} \right \}$ from the other camera views, we obtain $\Mat{f}_p^A$ and $\left \{ \Mat{\Bar{f}}_{j}^A \right \}$ and compute the between-camera matching distances using the Euclidean distance. We then rank all $\left \{ \Mat{\Bar{x}}_{j} \right \}$ in ascending order based on their distances given $\Mat{x}_i$ and use that to evaluate the identity of $\Mat{x}_p$.
\renewcommand{\arraystretch}{1.1}

\section{Experiments}\label{Experiment}
\textbf{Datasets and Evaluation Protocol} In this section, we show the effectiveness of our proposed algorithm through an extensive set of experiments across three well known person Re-ID datasets; (a) Market-1501~\cite{zheng2015scalable}, (b) DukeMTMC-reID (or DukeMTMC)~\cite{ristani2016MTMC}, (c) CUHK03~\cite{li2014deepreid} and (d) MSMT~\cite{wei2018person}. Market-1501 has $751/750$ train/test identity  split, and $32,668$ images in total. DukeMTMC-reID has $702/702$ train/test identity  split, and $36,411$ images in total. CUHK03 has $14,097$ images in total. In order to make the re-identification task more challenging on CUHK03, we use the $767/700$ train/test identity split~\cite{Re_rankingReID} instead of the $1367/100$ standard split. The train/test id split and the test protocol are shown in Table~\ref{table:DatasetSpecification}. The \textbf{MSMT17}~\cite{wei2018person} dataset consists of $126,441$ person images from $4,101$ identities, thus constituting the largest person Re-ID dataset at present. All person images are detected using a Faster R-CNN \cite{girshickICCV15fastrcnn}. This dataset is collected using $15$ different cameras; and the images were captured over $4$ different days experiencing different weather conditions during a month. The training set consists of $32,621$ images belonging to $1,041$ identities, whereas the test set contains $93,820$ images belonging to the remaining $3,060$ identities. The test set is further randomly divided into $11,659$ and $82,161$ images for query and gallery sets respectively. Both mean Average Precision (mAP) and Cumulative Matching Characteristic (CMC) metrics are used for measuring performance on these datasets. 

\begin{table}[t]
\caption{\small{The details of evaluated datasets. \emph{dis} refers to the distractor images of the DukeMTMC-reID dataset. TS, SQ, MQ and SS stand for Test Setting, Single Query, Multiple Query and Single Shot, respectively.}}\label{table:DatasetSpecification}
\centering
\scalebox{0.82}{
\begin{tabular}{|c|c|c|c|c|c|}
\hline
{Dataset} & {Images}  & {IDs} & {Train} & {Test} & {TS}\\ \hline
{Market1501} & {32,668} & {1501} & {751} & {750} & {SQ/MQ} \\ \hline
{DukeMTMC-reID} & \scalebox{0.9}{36,411} & \begin{tabular}[c]{@{}c@{}}\scalebox{0.9}{1404}\\ +\\ {408} \emph{{dis}} \end{tabular} & {702} & {702} & {SQ} \\ \hline
CUHK03 & 14,097 & 1467 & 767 & 700 & SS\\ \hline
MSMT17 & 126,441 & 4,101 & 1,041 & 3,060 & SQ \\ \hline
\end{tabular}
}
\end{table}

\begin{table}[h]
    \caption{Comparison results on Market-1501~\cite{zheng2015scalable} dataset.}
    \label{tab:market}
    \scalebox{0.76}{
        \begin{tabular}{|c|c|c|c|c|c|c|c|}
        \hline
        Method & SVDNet~\cite{SunYifan2017ICCVSVDNet} & MHAN~\cite{chen2019mixed}
        & Dare~\cite{Wang_2018_CVPR} & AOS~\cite{huang2018adversarially} & MLFN~\cite{MLFN} & SGGNN~\cite{Shen_2018_ECCV} \\ \hline
        mAP & 62.1 & 85.0 & 69.9 & 70.4 & 74.3 & 82.8 \\ \hline
        R1  & 82.3 & 95.1 & 86.0 & 86.5 & 90.0 & 92.3 \\ \hline \hline
        
        Method  & IANet~\cite{IANet} & PCB~\cite{sun2018beyond} & MSCAN~\cite{LiDangwei2017CVPRLearningDeepContextAwarefeaturesforReID} & JLML~\cite{LiDeepJoint} & PBR~\cite{Suh_2018_ECCV} & MGCAM~\cite{SongChunfeng2018CVPRMaskGuidedContrastiveAttentionModelforReID}  \\ \hline
        mAP & 83.1 & 81.6 & 57.5 & 65.5 & 76.0 & 74.3 \\ \hline
        R1  & 94.4 & 93.1 & 80.3 & 85.1 & 90.2 & 83.8 \\ \hline \hline
        
        Method & AANet~\cite{AANet_A} & HPN~\cite{LiuXihui2017ICCVHydraPlusNetforReID} & DKPM~\cite{Shen_2018_CVPR} & DuATM~\cite{DuATM} & Mancs~\cite{Mancs} & HA-CNN~\cite{LiWei2018CVPRHarmoniousAttentionNetforReID} \\ \hline
        mAP    & 83.4 & - &  75.3 & 76.6 & 82.3 & 75.7 \\ \hline
        R1     & 93.9 & 76.9 & 90.1 & 91.4 & 93.1 & 91.2 \\ \hline \hline
        
        Method & CASN~\cite{zheng2019re} & CAR~\cite{zhou2019discriminative} & OSNet~\cite{zhou2019omni} & DGNet~\cite{zheng2019joint} & CAMA~\cite{yang2019towards} & CCAN (Ours) \\ \hline
        mAP    & 82.8 & 84.7 & 84.9 & \textbf{\textcolor{red}{86.0}} & 84.5 & \textbf{87.0}\\ \hline
        R1     & 94.4 & \textbf{\textcolor{red}{96.1}} & 94.8 & 94.8 & 94.7 & \textbf{94.6} \\ \hline
\end{tabular}}
\end{table}

\begin{table}[t]
    \caption{Comparison results on DukeMTMC~\cite{ristani2016MTMC} dataset.}
    \label{tab:duke}
    \scalebox{0.78}{
    \begin{tabular}{|c|c|c|c|c|c|c|c|}
    \hline
    Method & SVDNet~\cite{SunYifan2017ICCVSVDNet} & IDE~\cite{zheng2016person}
    & Dare~\cite{Wang_2018_CVPR} & AOS~\cite{huang2018adversarially} & MLFN~\cite{MLFN} & SGGNN~\cite{Shen_2018_ECCV} \\ \hline
    mAP    & 56.8 & 64.2 & 56.3 & 62.1 & 62.8 & 68.2 \\ \hline
    R1     & 76.7 & 80.1 & 74.5 & 79.2 & 81.0 & 81.1 \\ \hline \hline
    
    Method  & IANet~\cite{IANet} & PCB~\cite{sun2018beyond} & MSCAN~\cite{LiDangwei2017CVPRLearningDeepContextAwarefeaturesforReID} & JLML~\cite{LiDeepJoint} & PBR~\cite{Suh_2018_ECCV} & MGCAM~\cite{SongChunfeng2018CVPRMaskGuidedContrastiveAttentionModelforReID}  \\ \hline
    mAP    & 73.4 & 69.7 & - & 56.4 & 64.2 & - \\ \hline
    R1     & 87.1 & 83.9 & - & 73.3 & 82.1 & - \\ \hline \hline
    
    Method & AANet~\cite{AANet_A} & HPN~\cite{LiuXihui2017ICCVHydraPlusNetforReID} & DKPM~\cite{Shen_2018_CVPR} & DuATM~\cite{DuATM} & Mancs~\cite{Mancs} & HA-CNN~\cite{LiWei2018CVPRHarmoniousAttentionNetforReID} \\ \hline
    mAP    & 74.3 & - & 63.2 & 64.6 & 71.8 & 63.8 \\ \hline
    R1     & 87.7 & - & 80.3 & 81.8 & 84.9 & 80.5 \\ \hline \hline
    
    Method & CASN~\cite{zheng2019re} & CAR~\cite{zhou2019discriminative} & OSNet~\cite{zhou2019omni} & DGNet~\cite{zheng2019joint} & CAMA~\cite{yang2019towards} & CCAN (Ours) \\ \hline
    mAP    & 73.7 & 73.1 & 73.5 & \textbf{\textcolor{red}{74.8}} & 72.9 & \textbf{76.8} \\ \hline
    R1     & 87.7 & 86.3 & \textbf{\textcolor{red}{88.6}} & 86.6 & 85.8 & \textbf{87.2}\\ \hline

\end{tabular}}
\end{table}

\subsection{Implementation}\label{Implemetation}
Our CCAN model is implemented in PyTorch \cite{paszke2017automatic}.
We use GoogLeNet-V1~\cite{szegedy2015going} with Batch Normalization~\cite{ioffe2015batch} pretrained on Imagenet~\cite{russakovsky2015imagenet} as our backbone architecture. The dimensionality of the output feature maps of the global branch (\emph{i.e}, $\mathrm{I}_1^G$, $\mathrm{I}_2^G$ and $\mathrm{I}_3^G$) is fixed to $480$, $832$, and $1024$ respectively. Similarly, in the local branch, the dimensionality of the output feature maps of $\mathrm{I}^L_{2,s}$ and $\mathrm{I}^L_{3,s}$ is set to $832$, and $1024$ for every $s$ respectively.  The embedding dimension $d$ and the number of local parts~(\emph{i.e} $k_p$) are set to $1024$ and $4$ across all the four datasets. None of the Inception and FC layers share weights between each other. The ADAM optimizer~\cite{kingma2014adam} is used to train the model, with the two moment terms ($\beta_1, \beta _2$), and the weight decay set to ($0.9$, $0.99$) and $1 \times 10^{-4}$, respectively. The learning rate is initially set to $5 \times 10^{-4}$ for Market-1501 and DukeMTMC-reID; and $1 \times 10^{-3}$ for CUHK03 in both the labeled and detected settings; which is fixed for the first $150$ epochs and decayed by a factor of $0.1$ after every $50$ epochs thereafter. The batch size is set to $64$ of $16$ identities with $4$ images per identity in all the datasets. The smoothing parameter $\epsilon$ of LSR is $0.1$. The margin $\tau$ for the triplet loss (Refer to Eqn.~\ref{eqn:triplet}) is set to $1$ for Market-1501 and DukeMTMC-reID, and $1.5$ for CUHK03 in both the dataset settings. The training images are first resized to $288 \times 144$ and then randomly cropped to $256 \times 128$, followed by a random horizontal flip. Following the protocol of~\cite{Mancs}, we apply random erasing \cite{zhong2017random} after the $\nth{50}$ epoch. However, during the test phase, the images are resized to $256 \times 128$ without any such data-augmentation techniques. We report the results after $200$ epochs of training.

\subsection{Comparison to State-of-the-Art Methods\protect\footnote{We report our results in \textbf{bold}, while we use \textbf{\textcolor{red}{red}} to report the best previous results obtained so far.}}

\subsection*{Evaluation on Market-1501} We have evaluated against a number of recently proposed methods with, or without, the use of attention modules. Table~\ref{tab:market} clearly shows the superior performance of CCAN against all the other methods in terms of mAP and Rank-1 accuracies on the Market-1501 dataset. More specifically, CCAN improves over the current state-of-the-art AANet by a prominent margin in the single query setting. We also outperform hard and soft attention based HA-CNN by $11.3/3.4\%$ with respect to mAP and Rank-1 respectively in the single query setting.

\subsection*{Evaluation on DukeMTMC-reID} We further evaluated our proposed CCAN on the DukeMTMC-reID~\cite{ristani2016MTMC} dataset. More variations in resolution and viewpoints due to wider camera views, and more complex environmental layout make DukeMTMC-reID more challenging compared to the Market-1501 dataset for the task of Re-ID. Table~\ref{tab:duke}~
shows that CCAN again outperforms almost all the baseline algorithms, except AANet in terms of Rank-1. However, we achieve higher mAP by a significant margin. We also outperform hard and soft attention based HA-CNN by $13.0/6.7\%$ with respect to mAP and Rank-1 respectively.

\begin{table}[h]
    \centering
    \caption{Comparison results on CUHK03 dataset in both the \textbf{Labeled} and the \textbf{Detected} settings.}
    \label{tab:cuhk_03}
    \scalebox{1}{
    \begin{tabular}{|c|c|c|c|c|}
        \hline
         & \multicolumn{2}{c|}{Labeled} & \multicolumn{2}{c|}{Detected} \\ \hline
        Measure (\%) & mAP & R1 & mAP & R1 \\ \hline
        MLFN \cite{MLFN} & 49.2 & 54.7 & 47.8 & 52.8 \\ 
        IDE~\cite{zheng2016person} & 48.5 & 52.9 & 46.3 & 50.4 \\
        AOS  \cite{huang2018adversarially} & - & - & 47.1 & 43.4 \\ 
        Dare~(De)~\cite{Wang_2018_CVPR} & 52.2 &  56.4 & 50.1 & 54.3 \\
        PCB~\cite{sun2018beyond} & 56.8 & 61.9 & 54.4 & 60.6 \\
        SVDNet~\cite{SunYifan2017ICCVSVDNet} & - & - & 37.3 & 41.5 \\ 
        MGCAM\cite{SongChunfeng2018CVPRMaskGuidedContrastiveAttentionModelforReID} & 50.2 & 50.1  & 46.9 & 46.7 \\
        Mancs\cite{Mancs} & 63.9 & 69.0 & 60.5 & 65.5 \\ 
        
        HA-CNN\cite{LiWei2018CVPRHarmoniousAttentionNetforReID} & 41.0 & 44.4 & 38.6 & 41.7 \\ 
        CAMA~\cite{yang2019towards} & - & - & 64.2 & 66.6 \\
        OSNet~\cite{zhou2019omni} & - & - & \textbf{\textcolor{red}{67.8}} & \textbf{\textcolor{red}{72.3}}\\
        CASN~\cite{zheng2019re} & \textbf{\textcolor{red}{68.0}} & \textbf{\textcolor{red}{73.7}} & 64.4 & 71.5 \\ 
        
        \hline
        CCAN (Ours) & \textbf{72.9} & \textbf{75.2} & \textbf{70.7} & \textbf{73.0} \\
        \hline
\end{tabular}}
\end{table}

\subsection*{Evaluation on CUHK03} We have also evaluated CCAN on both the manually \emph{labeled} and \emph{detected} person bounding boxes versions of CUHK03. The $767/700$ split results in a small training set with only $7365$ images against $12,936/16,522$ training images in Market-1501/DukeMTMC-reID datasets respectively. Even with such a constrained training setting, Table~\ref{tab:cuhk_03} clearly shows that notable improvement for CCAN against the baseline methods, including the current state-of-the-art Mancs, in both the labeled and detected settings. Furthermore, we also outperform HA-CNN by $31.9/30.8\%$ and $32.1/31.3\%$ in terms of mAP and Rank-1 in both the settings respectively.

\subsection*{Evaluation on MSMT17}

Table \ref{tab:MSMT} shows the result of our proposed CCAN when trained and evaluated on the new challenging MSMT17~\cite{wei2018person} dataset. As can be seen, CCAN achieves a significant performance gain with regards to mAP and Rank-1 over all the baseline algorithms. Specifically, CCAN outperforms the current state-of-the-art algorithm on MSMT, \emph{i.e.} Glad~\cite{glad}, by $19.6/14.9\%$ in terms of mAP and Rank-1 respectively.

\begin{table}[h]
    \centering
    \caption{Comparison results on MSMT17 dataset.}
    \label{tab:MSMT}
    \scalebox{1.1}{
    \begin{tabular}{|c|c|c|c|c|}
        \hline
        Model& mAP & R-1 & R-5 & R-10\\ \hline
        GLNet~\cite{szegedy2015going}  & 23.0 & 47.6 & 65.0 & 71.8\\
        PDC~\cite{SuChi2017ICCVPoseDrivenDeepModelforReID} & 29.7 & 58.0 & 73.6 & 79.4\\
        Glad~\cite{glad} & 34.0 & 61.4 & 76.8 & 81.6 \\
        PCB~\cite{sun2018beyond} & 40.4 & 68.2 & - & - \\
        OSNet~\cite{zhou2019omni} & \textbf{\textcolor{red}{52.9}} & \textbf{\textcolor{red}{78.7}} & - & - \\
        IANet~\cite{IANet} & 46.8 & 75.5 & - & - \\
        DGNet~\cite{zheng2019joint} & 52.3 & 77.2 & - & - \\ 
        \hline
        CCAN (Ours) & \textbf{53.6} & \textbf{76.3} & \textbf{86.9} & \textbf{90.2} \\
        \hline
    \end{tabular}}
\end{table}

These results, on all the four challenging datasets mentioned above, clearly demonstrate and validate our proposed approach of cross-correlation based joint attention and discriminative feature learning for person Re-ID. CCAN outperforms all the current methods that rely only on hard, soft, or a combination of these two types of attention.
\section{Ablation Study}
In this section, we undertake a detailed study of the various aspects of our proposed CCAN framework. 
\label{results:AS}

\begin{figure*}[t]
	\centering
	\subfigure[]{\includegraphics[width=0.45\linewidth,keepaspectratio]{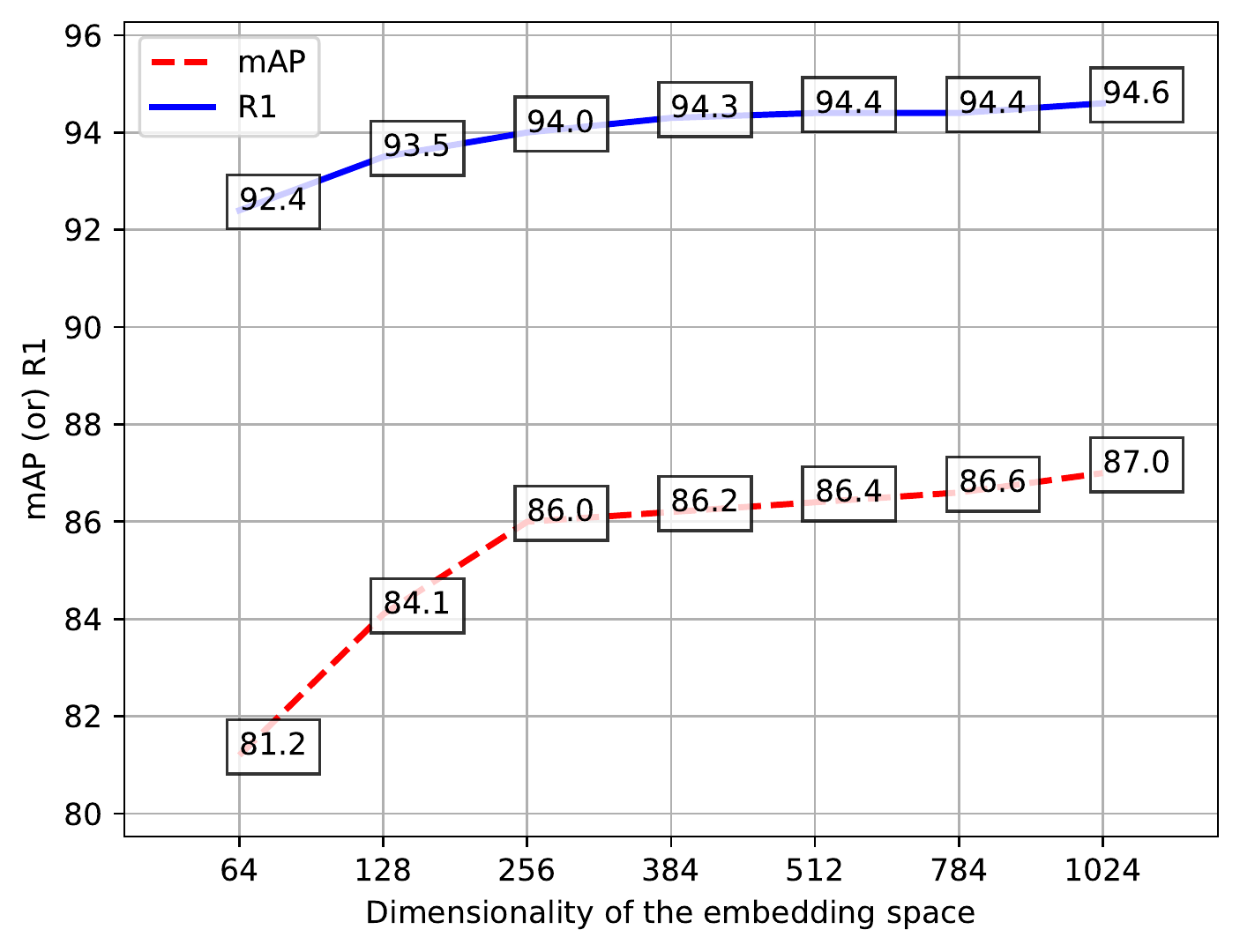}\label{fig:dimensionality}}%
	\hspace{0.5cm}
	\hfil
	\subfigure[]{\includegraphics[width=0.45\linewidth,keepaspectratio]{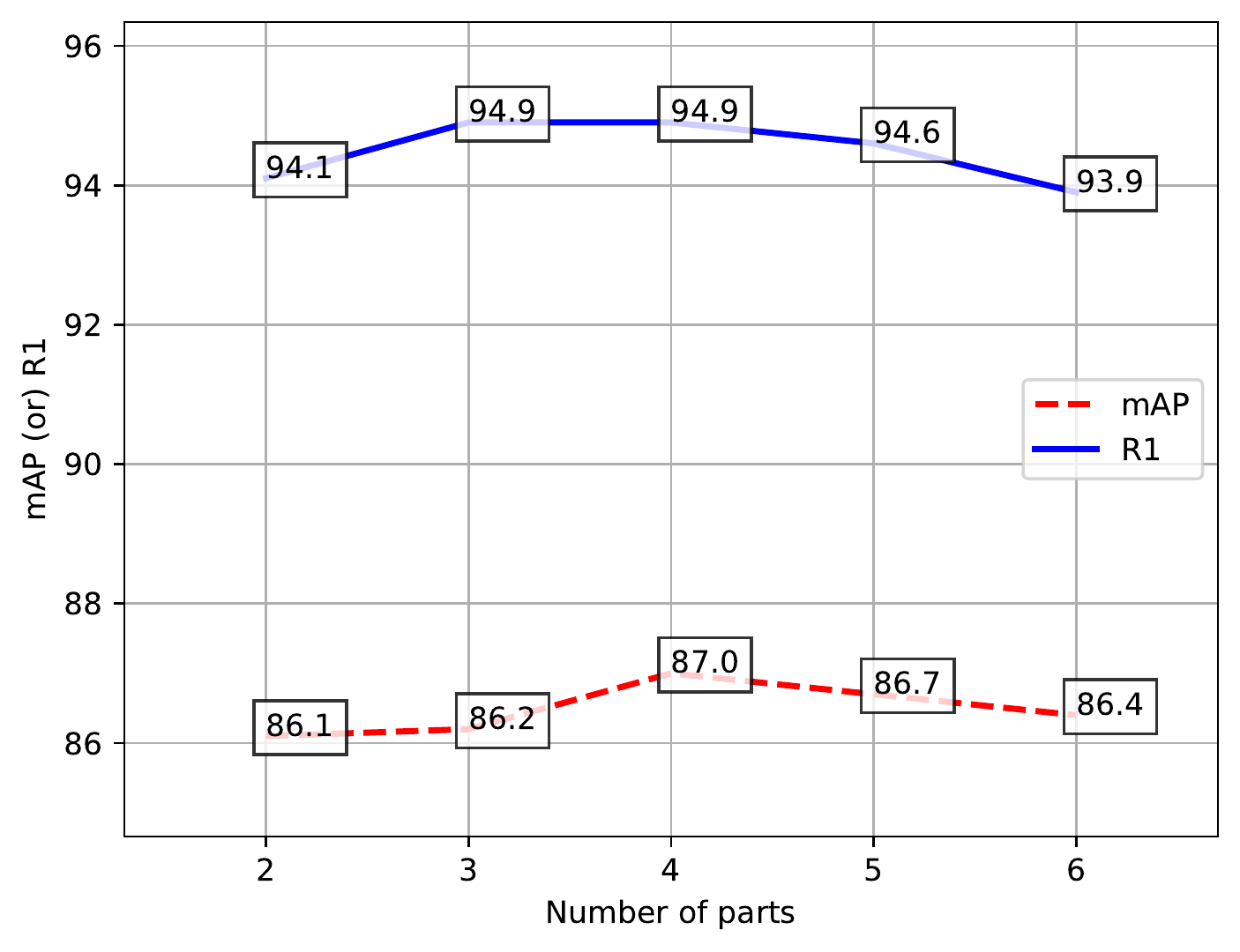}\label{fig:parts}}
	\caption{Ablation study of the (a) dimensionality of the embedding space (\emph{i.e.} $d$) and (b) number of body parts (\emph{i.e.} $k_p$). Both the experiments were conducted on Market-1501 in the Single Query setting. }
 \end{figure*} 

\subsection{Dimensionality of the embedding space.}
We first evaluate CCAN for different values of $d$ on the Market-1501~\cite{zheng2015scalable} dataset. As observed in Fig.~\ref{fig:dimensionality}, both mAP and R1 continue to increase as $d$ is increased from $128$ to $1024$, with the highest values obtained when $d$ is set to $1024$. Based on this experimental study, we decided to choose $1024$ as the embedding dimension $d$ for all the experiments. It is to be noted that even with a smaller $d$ (such as $256$), we still outperform all baseline algorithms (Refer to Table~\ref{tab:market_duke}). This clearly shows that CCAN is able to learn discriminative features and achieve state-of-the-art results for a large range of $d$.

\subsection{Number of body parts} We further evaluated the effect of various number of parts, \emph{i.e.}, $k_p$ in CCAN. Fig.~\ref{fig:parts} provides a detailed overview of the following evaluation for five different values of $k_p$. It can be seen that CCAN performs the best when $k_p$ is set to $4$, thereby suggesting that CCAN is able to detect and focus on the $4$ distinct regions of the input person image; namely (a) head-shoulder, (b) upper-body, (c) thighs, and (d) crus-foot. It should also be noted that even with $2$ different parts, CCAN is able to achieve competitive results against several baseline algorithms. This indeed demonstrates that CCAN is successful in exploiting the complementary nature of the learnt CCA attention modules even when lesser number of parts are specified. Based on this, in all the subsequent experiments, we have fixed the dimensionality of the embedding space (\emph{i.e.} $d$) to $1024$ and the number of parts (\emph{i.e.} $k_p$) to $4$.

\subsection{Importance of various attention modules}

We perform an ablation study in order to study the importance of various attention modules in CCAN. The results, evaluated on Market1501 dataset~\cite{zheng2015scalable} single query setting, are shown in Table~\ref{tab:ablation_study}. The following critical insights are observed : \textbf{(a)} The performance of the global branch $G$ (Id = $1$) and the local branch $L$ (Id = $2$) by itself reads as $81.7\%$ and $79.5\%$ mAP respectively. \textbf{(b)} Though combination of $G$ and $L$ helps (Id = $3$), incorporating only $\mathrm{SS}^G_1$ along $G$ (Id = $4$) leads to almost similar performance. \textbf{(c)} Furthermore, Id=$3$ and $5$ show the importance of adding a CCA module, \emph{i.e} $\mathrm{CC}^L_2$, along $L$.\textbf{(d)} Finally CCAN improves over Id=$6$ with the addition of a $\mathrm{SS}^G_1$ along $G$ (Refer to Fig~\ref{fig:CCANstrucutre}). This indeed verifies the joint interactive learning of the attention modules and feature extractors to obtain a discriminative embedding space for the person images. It is to be noted that in all our experiments, we have kept the final structure of CCAN fixed across all the datasets, suggesting a novel and rich architecture for the task of Re-ID that generalises well.

\begin{table}[h]
    \centering
    \caption{Study of the importance of various attention modules on Market-1501 dataset.}
    \label{tab:ablation_study}
    \scalebox{0.85}{
    \begin{tabular}{|c|c|c|c|c|c|c|}
            \hline
            Id & 1 & 2 & 3 & 4 & 5 & 6 \\ \hline
            Setting & G & L & G+L & G+$\mathrm{SS}^G_1$ & G+L+$\mathrm{CC}^L_2$ & \textbf{CCAN} \\ \hline
            mAP & 81.7 & 79.5 & 83.6 & 83.3 & 85.6 & \textbf{87.0} \\ \hline
            R1 & 92.7 & 92.1 & 93.3 & 92.9 & 94.3 & \textbf{94.6} \\ \hline
    \end{tabular}}
\end{table}

\section{Conclusions}\label{Conclusion}
In this paper, we propose a new attention module, called \emph{Cross-Correlated Attention} (CCA), which aims to improve the information gain by learning to focus on the correlated regions of the input image. We incorporate CCA into a novel deep attention architecture that we name \emph{Cross-Correlated Attention Network} (CCAN) to achieve state-of-the-art results on three challenging datasets by utilizing the complementary nature of the attention mechanisms. In contrast to most existing attention based Re-ID models that use constrained attention learning algorithms, CCAN is capable of exploring and exploiting correlated interaction among the attention modules to locate and focus on the discriminative regions of the input person image without the need of any part (or pose) based estimator or detector network in a unified end-to-end CNN architecture. In the future, we plan to design and incorporate attention-diversity loss into CCAN to obtain further improvements and better focused attention maps. We also plan to study the effects of augmenting CCAN with additional part/pose estimation or detection networks in the future.



\bibliographystyle{elsarticle-num} 
\bibliography{main.bib}





% \end{thebibliography}
% \doclicenseThis
\end{document}